\documentclass[a4paper,10pt]{article}
\usepackage{aaai}
\usepackage{times}
\usepackage{helvet}
\usepackage{courier}
\title{Lazy Explanation-Based Approximation for Probabilistic Logic Programming}
\author{Joris Renkens \and Angelika Kimmig \and Luc De Raedt}

\begin{document}

\setcounter{secnumdepth}{1}
\maketitle

\begin{abstract}
We introduce a lazy approach to the explanation-based approximation of probabilistic logic programs. It uses only the most significant part of the program when searching for explanations. The result is a fast and anytime approximate inference algorithm which returns hard lower and upper bounds on the exact probability. We experimentally show that this method outperforms state-of-the-art approximate inference.
\end{abstract}

\section{Introduction}
Probabilistic logic programming (PLP) languages extend logical languages with probabilities. Examples of such languages are PRISM \cite{Sato95}, ICL \cite{Poole2008}, LPADs \cite{Vennekens2004} and ProbLog \cite{DeRaedt2007}. 

A typical inference task in these languages is calculating the probability of a query. It is often computed by transforming the program into a weighted propositional formula and subsequently calculating the weighted model count (WMC) \cite{Chavira2008} of the formula. When formulae become large, computing the WMC becomes prohibitively expensive and approximating methods are needed. 

Explanation-based inference methods \cite{Kimmig2008} approximate the probability by constructing a smaller propositional formula on which the WMC is calculated. Earlier work \cite{Renkens2014} developed an explanation-based algorithm in the presence of negation. The downside of this approach is that it uses the entire program when constructing the formula. We introduce a lazy approach that avoids this, which enables it to get results, even when the program is very large. The result is a fast anytime approximate inference algorithm which returns hard lower and upper bounds on the exact probability.

In the remainder of the paper we will give some background on PLP and explanation-based approximation in Section~\ref{background}, explain the intuition behind the lazy approach in Section~\ref{lazy}, provide experimental results in Section~\ref{experiments} and conclude in Section~\ref{conclusions}.

\section{Probabilistic Logic Programming}
\label{background}
Most PLP languages are based on Sato's distribution semantics \cite{Sato95}. In this paper, we will use ProbLog as the example language. We will limit ourselves to ground programs but the approach is easily extended to non-ground programs. 

A ProbLog program consists of two parts: a set of independent \textbf{probabilistic facts} $\mathcal{F}$ and a set of \textbf{rules} $\mathcal{R}$. Each probabilistic fact has the form {\small\verb=p::atom.=} stating that {\small\verb=atom=} is true with probability {\small\verb=p=}. Each rule is a logic programming rule of the form {\small\verb= head :- body.=} where {\small\verb=head=} is an atom and {\small\verb=body=} is a conjunction of literals. Atoms that appear as the head of a rule are called \textbf{defined atoms} and we assume the set of defined atoms and probabilistic facts to be disjoint.

A full interpretation of the probabilistic facts is called a \textbf{possible world} $\mathcal{W}$. The probability of a possible world is defined as the product of the probabilities of the literals in the world.  Together with the rules $\mathcal{R}$, each possible world assigns unique truth values to the defined atoms. When $\mathcal{W} \cup \mathcal{R} \models q$ holds, we say that $q$ is true in the possible world $\mathcal{W}$. When calculating the probability of a query $q$, only the worlds in which the query is true are taken into account: $P(q) = \sum_{\mathcal{W} \cup \mathcal{R} \models q}\prod_{l \in \mathcal{W}} P(l)$

An \textbf{explanation} $\mathcal{E}$ for a query is a partial interpretation of the probabilistic facts. It specifies a set of possible worlds in which the query is true. Since the probabilistic facts are independent, the probability of an explanation can be calculated as: $P(\mathcal{E}) = \prod_{l \in \mathcal{E}} P(l)$

Renkens et al. (2014) use explanations to construct formulae for the upper as well as lower bound of a query $q$. When we write an explanation as a conjunction of literals, it is clearly a formula for the lower bound of the query. It captures a set of possible worlds in which $q$ is true, but not all of them. This lower bound can be improved by taking the disjunction of multiple explanations. Similarly, a disjunction of explanation for $\lnot q$ leads to an upper bound on the probability. Renkens et al. (2014) search optimal (highest probability) explanations for $q$ and $\lnot q$, by transformation to a weighted partial MAX-SAT problem, until time runs out.

\newpage

\section{Lazy search}
\label{lazy}
The approximation algorithm used in this paper is identical to the one in \cite{Renkens2014} except for one key part. In contrast to Renkens et al. (2014), we only use part of the program (the lazy program) when searching explanations. This can lead to efficiency gains when the original program is big. We will show the intuition behind this approach on a toy example. It defines paths in a network. {\small\verb=p(x,y)=} and {\small\verb=e(x,y)=} state that there is a path and edge respectively, between nodes {\small\verb=x=} and {\small\verb=y=}. We show the rules and facts for query {\small\verb=p(1,4)=}, for which we will search explanations:

\begin{itemize}
 \item[] {\small\textbf{Rules:} $\{\verb=p(1,4):-e(1,2),p(2,4)=;\verb=p(2,4):-e(2,4)=;\\\verb=p(1,4):-e(1,3),p(3,4)=;\verb=p(3,4):-e(3,4)=\}$}
 \item[] {\small\textbf{Facts:} $\{\verb=0.8::e(1,2)=;\verb=0.1::e(1,3)=;\verb=0.5::e(2,4)=;\\\verb=0.4::e(3,4)=\}$}
\end{itemize}

When the search is started, we add all probabilistic facts to the lazy program. However, we add none of the rules to the program and instead add for each head of a rule, a weighted fact {\small\verb=(1;1)::head=}. The fact \verb=head= will have a weight equal to one both when it is true as well as false. This means that if \verb=head= receives a truth value in an explanation, it always multiplies the probability of the explanation with one. We call \verb=head= unexpanded.

\begin{itemize}
 \item[] {\small\textbf{Rules:} $\emptyset$}
 \item[] {\small\textbf{Facts:} $\{\verb=0.8::e(1,2)=;\verb=0.1::e(1,3)=;\verb=0.5::e(2,4)=;\\\verb=0.4::e(3,4)=;\verb=(1;1)::p(1,4)=;\verb=(1;1)::p(2,4)=;\\\verb=(1;1)::p(3,4)=\}$}
\end{itemize}

Subsequently, the optimal explanation for {\small\verb=p(1,4)=} is searched in the lazy program. This is done in the same way as in Renkens et al. (2014) but is easier since the lazy program is smaller. The resulting explanation is {\small$\{\verb=p(1,4)=\}$}. When the optimal explanation in the lazy program contains unexpanded heads, they are replaced by their rules and the search for the optimal explanation is repeated. The program for the next iteration is:

\begin{itemize}
 \item[] {\small\textbf{Rules:} $\{\verb=p(1,4):-e(1,2),p(2,4)=;\\\verb=p(1,4):-e(1,3),p(3,4)=\}$}
 \item[] {\small\textbf{Facts:} $\{\verb=0.8::e(1,2)=;\verb=0.1::e(1,3)=;\verb=0.5::e(2,4)=;\\\verb=0.4::e(3,4)=;\verb=(1;1)::p(2,4)=;\verb=(1;1)::p(3,4)=\}$}
\end{itemize}

Again the optimal explanation {\small$\{\verb=e(1,2)=;\verb=p(2,4)=\}$} is searched and {\small\verb=(1;1)::p(2,4)=} is replaced by its rules.

\begin{itemize}
 \item[] {\small\textbf{Rules:} $\{\verb=p(1,4):-e(1,2),p(2,4)=;\verb=p(2,4):-e(2,4)=;\\\verb=p(1,4):-e(1,3),p(3,4)=\}$}
 \item[] {\small\textbf{Facts:} $\{\verb=0.8::e(1,2)=;\verb=0.1::e(1,3)=;\verb=0.5::e(2,4)=;\\\verb=0.4::e(3,4)=;\verb=(1;1)::p(3,4)=\}$}
\end{itemize}

Now, the optimal explanation {\small$\{\verb=e(1,2)=;\verb=e(2,4)=\}$} does not contain any unexpanded heads. In general, optimal explanations without unexpanded heads are also optimal explanations in the original program. Any explanation in the lazy program, containing unexpanded heads, needs additional facts to make it an explanation in the original program. This can never increase its probability.

\section{Experiments}
\label{experiments}

We experimentally evaluate our approximate inference algorithm by comparing to the non-lazy approach \cite{Renkens2014} and a state-of-the-art forward reasoning approach, which we will call TP \cite{Vlasselaer2015}. We evaluate 500 queries on the biological network of \cite{Ourfali2007} with a timeout of 15 minutes per query. The results can be found in Table~\ref{results}. They show that the lazy approach outperforms both other approaches.

\begin{table}[h]
\centering
\begin{tabular}{l|c|c|c}

&\small non-lazy & \small TP & \small lazy \\ \hline
\small Almost Exact & \small 0 & \small 30 & \small 89\\ \hline
\small Tight Bound & \small 0 & \small 207 &\small 272\\ \hline
\small Loose Bound & \small 0 & \small 263 & \small 139\\ \hline
\small No Answer & \small 500 &\small  0 &\small 0
\end{tabular}
\caption{\small The number of queries for which the difference between upper and lower bound is $< 0.01$ (Almost Exact), in $[0.01,0.25)$ (Tight Bounds), in $[0.25,1)$ (Tight Bounds) and $=1$ (No Answer)}
\label{results}

\end{table}

\section{Conclusions}
\label{conclusions}
We have proposed a lazy approach to explanation-based approximation for probabilistic logic programs. This approach outperforms non-lazy explanation-based methods as well as other state-of-the-art approaches when programs are large. While this paper only discusses the ground case, all techniques can be extended to the non-ground case.

\small
\bibliographystyle{aaai}
\bibliography{/home/jorisr/Documents/papers/references.bib}

\end{document}